\documentclass[10pt,twocolumn,letterpaper]{article}

\usepackage{cvpr}
\usepackage{times}
\usepackage{epsfig}
\usepackage{graphicx}
\usepackage{amsmath}
\usepackage{amssymb}
\usepackage{multirow}

\usepackage[breaklinks=true,bookmarks=false,colorlinks,linkcolor=red]{hyperref}

\cvprfinalcopy 

\def\cvprPaperID{1868} 

\newcommand{\grad}{{\displaystyle \nabla}}

\newcommand\cspace{15}
\newcommand\cspaceadd{2}

\ifcvprfinal\fi
\begin{document}

\title{Semantic Image Matting}

\author{Yanan Sun\\
HKUST\\
{\tt\small now.syn@gmail.com}

\and
Chi-Keung Tang\\
HKUST\\
{\tt\small cktang@cs.ust.hk}

\and
Yu-Wing Tai\\
Kuaishou Technology\\
{\tt\small yuwing@gmail.com}

}

\maketitle

\begin{abstract}
Natural image matting separates the foreground from background in fractional occupancy which can be caused by highly transparent objects, complex foreground (e.g., net or tree), and/or objects containing very fine details (e.g., hairs). Although conventional matting formulation can be applied to all of the above cases, no previous work has attempted to reason the underlying causes of matting due to various foreground semantics.

We show how to obtain better alpha mattes by incorporating into our framework semantic classification of matting regions. Specifically, we consider and learn 20 classes of matting patterns, and propose to extend the conventional trimap to semantic trimap. The proposed semantic trimap can be obtained automatically through patch structure analysis within trimap regions. Meanwhile, we learn a multi-class discriminator to regularize the alpha prediction at semantic level, and content-sensitive weights to balance different regularization losses. Experiments on multiple benchmarks show that our method outperforms other methods and has achieved the most competitive
state-of-the-art performance. Finally, we contribute a large-scale Semantic Image Matting Dataset with careful consideration of data balancing across different semantic classes. Code and dataset are available at 
{\small \url{https://github.com/nowsyn/SIM}}.{\let\thefootnote\relax\footnotetext{{This work was done when Yanan Sun was a student intern at Kuaishou Technology. This work was supported by Kuaishou Technology and the Research Grant Council of the Hong Kong SAR under grant no. 16201420.}}}

\end{abstract}


\vspace{-10pt}
\section{Introduction}
\begin{figure}[t]
\centering 
\includegraphics[width=1.0\linewidth]{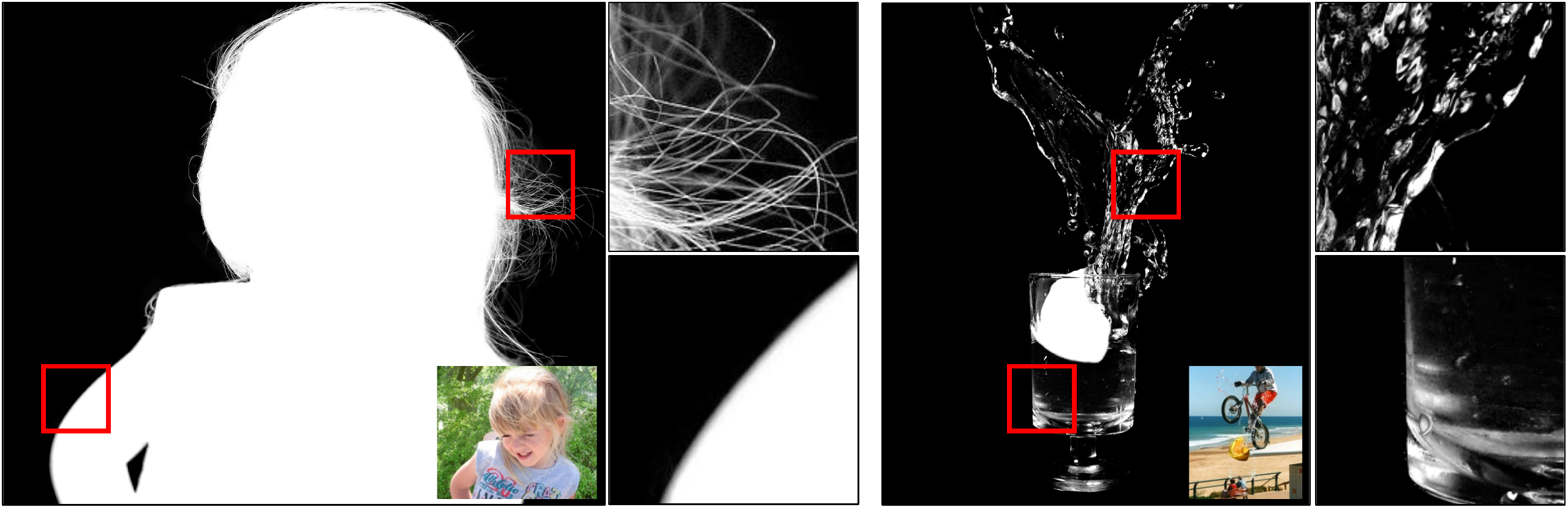} 
\caption{Alpha mattes can have different local shapes and patterns depending on the underlying foreground. Left shows a human matte with both soft hair and sharp body boundaries. Right shows a complex matte where the foreground  exhibits different degrees of transparency. Deep reasoning of object semantics in natural image matting can advance the state-of-the-art results.}
\label{fig:teaser_image}
\vspace{-14pt}
\end{figure}

The matting equation models an image $I$ as a linear combination of a foreground image $F$ and a background image $B$ modulated by an alpha matte $\alpha$:
\begin{equation}\label{eq:matting}
    I = \alpha F + (1-\alpha) B.
\end{equation} 
The natural image matting problem is to extract the alpha matte $\alpha$ from a given image, which has  a wide range of applications in image/video editing.
Typical foreground objects can belong to a great variety of categories, such as humans, furry animals, glass objects with transparent/semi-transparent regions, or objects with complex shapes such as net or tree, thus making this research problem still challenging, see Figure~\ref{fig:teaser_image}.

Natural image matting is an ill-posed problem, so most existing methods take both image and user-supplied clues (scribbles, trimaps, etc.) as input to solve the problem. Class-agnostic trimap comprising of foreground, background, and unknown regions is the most common choice among traditional and deep learning-based methods.

Traditional methods~\cite{Chuang-2001-CVPR-bayesianmatting,Feng-2016-ECCV-clustersampling,GastalOliveira-2010-CGF-SharedMatting,He-2011-cvpr-globalsampling,Ruzon-2000-cvpr-alphaestimation, Aksoy-2017-cvpr-ifm,bai-iccv-geodesicframework,Chen-2012-PAMI-knnmatting,grady2005random,levin2008closed,levin2008spectral}  mainly rely on low-level features related to image colors or structures. As traditional methods do not take into consideration any semantic information of foreground objects or background scenes, they fail easily in images where foreground pixels mingle deeply with background pixels. This issue has been addressed to a considerable extent with the introduction of deep neural networks. In~\cite{Aksoy-2018-tg-sss} images are segmented into soft transitional regions based on semantics. However, this method can easily fail in images with entangled colors or large transitional regions. To extract alpha mattes for general objects, deep learning-based methods~\cite{Cho-2016-mattingusingdeepcnn, Xu2017DeepIM, Lutz2018AlphaGANGA, cai2019disentangled, hou2019context, li2020natural} have contributed  various convolution network designs to greatly improve performance benefiting from high-level semantic representations. Recent works such as human matting and transparent object matting~\cite{Shen2016DeepAP, zhu2017fastdeepmattingportrait, Liu_2020_CVPR, chen2018tomnet} have focused on solving specific instances of the matting problem, leveraging prior domain knowledge, and contributing to excellent performance for these class-specific matting tasks.
However, these deep learning-based methods still utilize semantic information only at a data level and do not adequately consider the underlying cause of matting due to different object semantics. Given the unknown region surrounding a foreground object, different types of boundaries or patterns may exist. For example, a portrait usually has both fuzzy hair and sharp boundaries. 

In this paper, we propose to incorporate semantic classification of matting regions into our matting framework for extracting better alpha matte. Specifically, we first cluster 20 different matting classes based on the regional matting patterns. Our matting classes cover most typical matting scenarios for various foreground objects. Then, we extend the conventional class-agnostic trimap to semantic trimap, which consists of a 2D confidence map for each matting class in the unknown region. The automatically generated semantic trimap with RGB image is then fed into our framework as input. Meanwhile, we learn a multi-class discriminator for supervision, providing  regularization from a semantic level for alpha predictions. In particular, to improve prediction with semantics incorporation, we introduce gradient constraints with content-sensitive weights to balance different regularization losses. 

In summary, our main contributions are:
\begin{enumerate}
\vspace{-0.05in}
    \item We introduce semantics into the matting task and demonstrate how semantic information can be used to achieve the most competitive performance. To our knowledge, this is the {\em first} paper to consider semantic classification of matting patterns in a natural image matting framework.
    \vspace{-0.05in}
    \item Our main technical contributions include: the introduction of semantic trimap, the proposal of learnable content-sensitive weights, and the usage of multi-class discriminator to regularize the matting results.
    \vspace{-0.05in}
    \item We contribute the first large-scale class-balanced Semantic Image Matting Dataset covering a wide range of matting patterns to benefit future matting research. Our new dataset provides  new insight and  in-depth analysis to the performance of different matting algorithms on different matting classes.
\end{enumerate}

\section{Related Work}
We first briefly review representative works on general image matting. Then, we will discuss methods that are tailored for humans and other class-specific objects.

\vspace{4pt}
\noindent{\textbf{Natural Image Matting.}} Traditional methods on natural image matting are all based on the linear combination equation Eq.~\ref{eq:matting} except for~\cite{tai2007soft,kong2009transductive}. Since this is an ill-posed problem, traditional methods often make use of different priors. They can be further divided into sampling-based methods~\cite{Chuang-2001-CVPR-bayesianmatting,Feng-2016-ECCV-clustersampling,GastalOliveira-2010-CGF-SharedMatting,He-2011-cvpr-globalsampling,Ruzon-2000-cvpr-alphaestimation} and propagation-based methods~\cite{Aksoy-2017-cvpr-ifm,Aksoy-2018-tg-sss,bai-iccv-geodesicframework,Chen-2012-PAMI-knnmatting,grady2005random,levin2008closed,levin2008spectral} or a combination of both. Sampling-based methods assume color similarities between unknown regions and definitely foreground/background regions to estimate foreground and background colors followed by alpha matte estimation. Propagation-based methods utilize the color line model proposed by~\cite{levin2008closed} to estimate alpha by propagating its values from known regions to unknown regions. A comprehensive review on traditional matting methods can be found in~\cite{wang2008mattingsurvey,Cho-2016-mattingusingdeepcnn}.

For deep-learning based methods, Cho~\etal~\cite{Cho-2016-mattingusingdeepcnn} proposed to apply deep neural networks to fuse the results produced by closed-form matting~\cite{levin2008closed} and KNN matting~\cite{Chen-2012-PAMI-knnmatting}. Xu~\etal~\cite{Xu2017DeepIM} introduced the Composite-1K dataset and a two-stage encoder-decoder network for general object matting. Lutz~\etal~\cite{Lutz2018AlphaGANGA} proposed a generative adversarial network in the encoder-decoder framework to improve matting performance. Cai~\etal~\cite{cai2019disentangled} introduced trimap adaptation and applied multi-task learning to matting.
Hou~\etal~\cite{hou2019context} simultaneously estimated foreground and alpha matte  with local matting encoder and global context encoder.
Li~\etal~\cite{li2020natural} designed a guided contextual attention module to propagate high-level opacity information globally based on the learned low-level affinity.
Recently, a number of works~\cite{zhang2019latefusioncnn, BMSengupta20, qiao2020attention} have been proposed focusing on relaxing trimap input. Lastly, Forte and Piti\'{e}~\cite{forte2020fbamatting} introduced the FBA-net which simultaneously estimates foreground, background and alpha. See supplementary material for comparison with~\cite{forte2020fbamatting}\footnote{\cite{forte2020fbamatting} has not been published in recognized, peer-reviewed venues.}.

\vspace{4pt}
\noindent{\textbf{Human Matting.}}
Human matting was first introduced by Shen~\etal~\cite{Shen2016DeepAP} to extract alpha mattes of portrait images. Different from general object matting, human matting only considers hairs and sharp boundaries of human body. The authors formulated the solution as a two-step approach: human detection followed by human matting~\cite{zhu2017fastdeepmattingportrait}. Since foreground objects are already known to be human, the trimap can be eliminated~\cite{Chen2018SemanticHM,Liu_2020_CVPR}. Although these approaches give excellent performance for human subjects, they fall short of matting general objects. In addition, since they were trained  using portrait images, the performance would  significantly drop when the human detector fails.

\begin{figure*}[t]
\centering 
\includegraphics[width=1.0\linewidth]{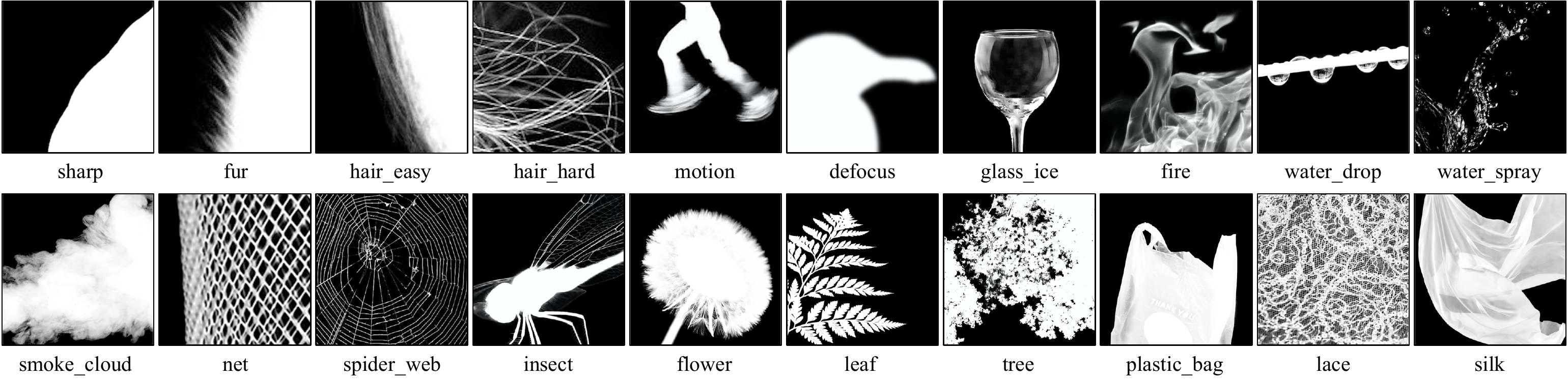} 
\caption{The 20 matting classes with high diversity in appearance across different classes.}
\label{fig:classes}
\vspace{-12pt}
\end{figure*}

\begin{figure}[t]
\centering 
\includegraphics[width=1.0\linewidth]{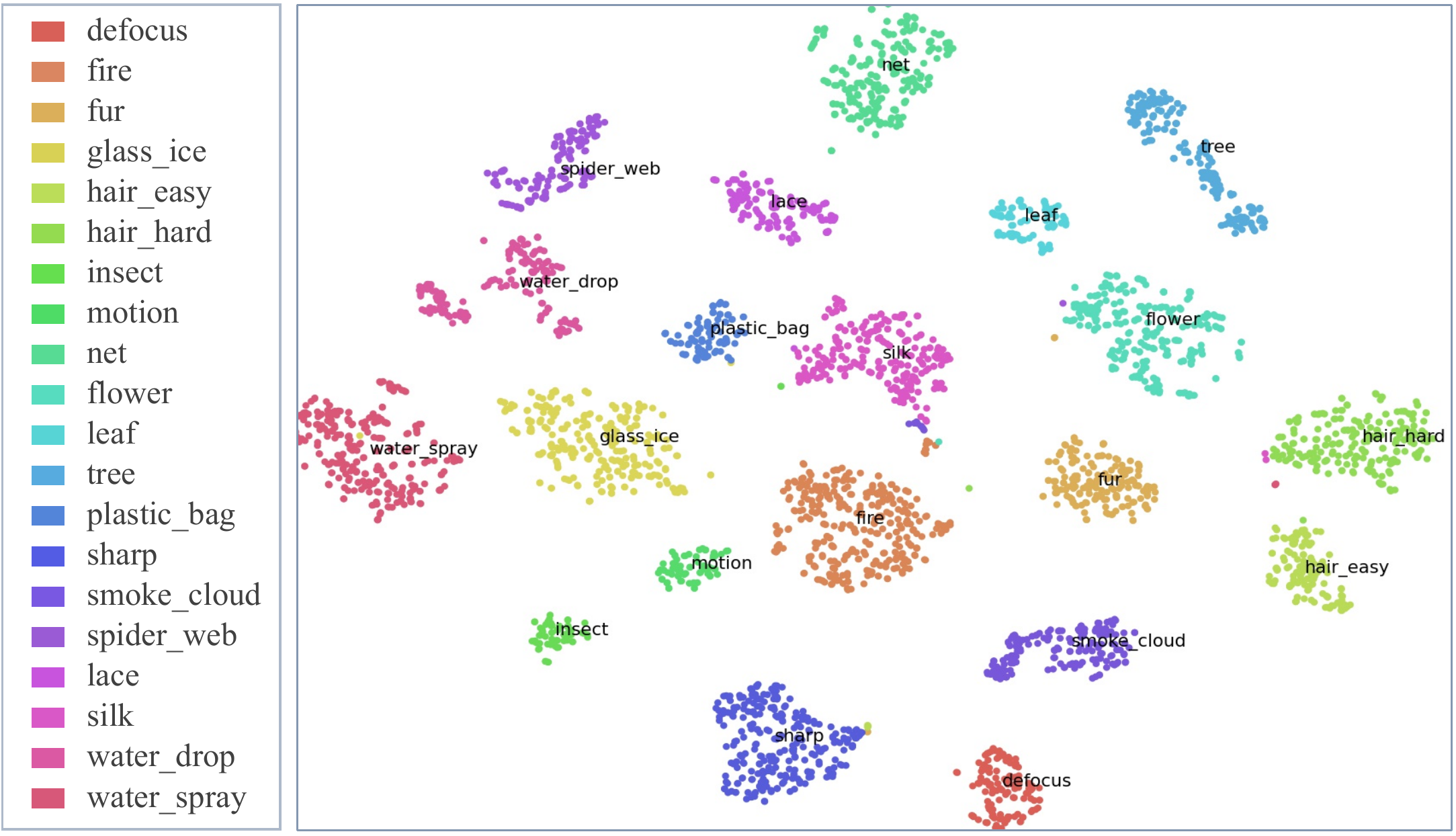} 
\caption{t-SNE visualizations of the class-specific features extracted from our discriminator (for its design see Method section).}
\label{fig:cluster}
\vspace{-12pt}
\end{figure}

\vspace{0.1in}

\vspace{2pt}
\noindent{\textbf{Class-Specific Matting.}}
For works  targeting at other specific classes of matting, Lin~\etal~\cite{lin2011motion} introduced motion regularization for matting  motion blurred moving objects. Köhler~\etal~\cite{kohler2013improving} proposed to separate motion blurred foreground through explicit modeling of object motion. Amin~\etal~\cite{amin2019hybrid} applied image matting to segment out-of-focus regions. Yeung~\etal~\cite{yeung2011tog} proposed attenuation-refraction matte to mask out and re-composite transparent object with refraction consideration. Chen~\etal~\cite{chen2018tomnet} proposed a multi-scale encoder-decoder network to estimate the refractive flow for transparent object matting.

In summary, traditional matting methods rely on low-level image cues for solving the matting equation, while more recent methods do not adequately consider or provide a principled approach to semantic image matting, or are limited to one or few classes such as humans.

\section{Dataset}
Although there exist several large-scale matting datasets for training deep neural networks,  they exhibit severe biases with respect to our {\em matting classes}.

\vspace{4pt}
\noindent{\textbf{Matting Classes.}} We study all existing publicly available matting datasets. Alpha mattes can exhibit diverse patterns. 
We propose 20 matting classes with distinct pattern as shown in Figure~\ref{fig:classes}.
These 20 classes cover almost all of the matting scenarios  in existing matting datasets. While there exist numerous matting cases in the real world, most of them can be categorized into these 20 classes while the rest are quite rare.
Figure~\ref{fig:cluster} shows the t-SNE visualizations of the 20 classes. 

Based on these classes, we keep part of foreground objects in Adobe Image Matting Dataset~\cite{Xu2017DeepIM}. Additionally, we collect sufficient clean images covering objects with corresponding patterns, and carefully extract their alpha matte with Photoshop. Finally, our Semantic Image Matting Dataset consists of 20 classes with 726 training foregrounds and 89 testing foregrounds. 

To generate the composited training set, we combine training foregrounds with randomly selected background images from COCO~\cite{lin2014microsoft} dataset in an online manner during training. For the testing set, we follow~\cite{Xu2017DeepIM} and composite each testing foreground with 10 background images from PASCAL VOC~\cite{Everingham10} dataset.

\begin{figure}[t]
\hspace{-0.2in}
\centering
\includegraphics[width=1\linewidth]{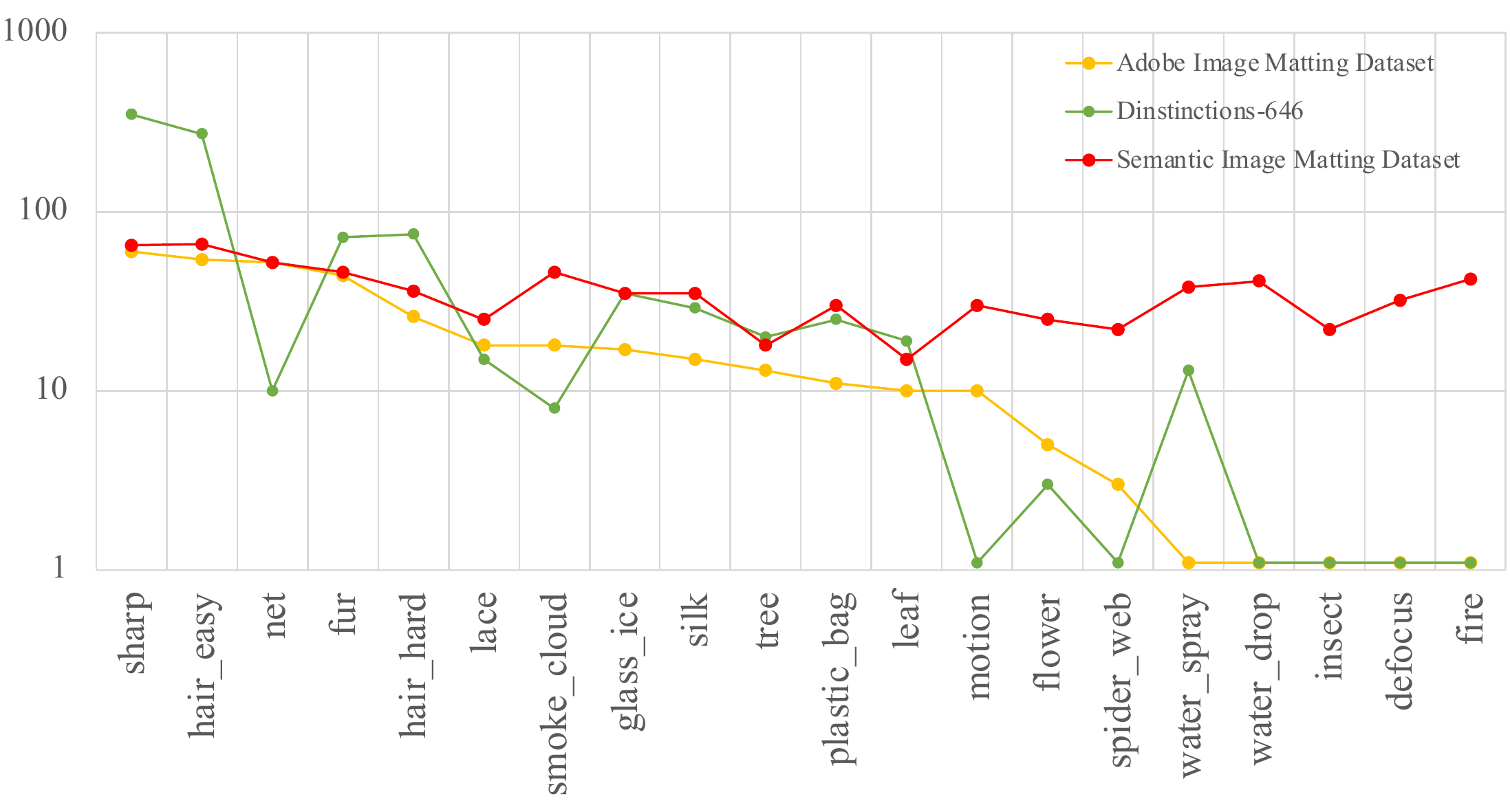} 
\caption{Class distribution of three matting datasets: Adobe Image Matting Dataset~\cite{Xu2017DeepIM}, Distinctions-646~\cite{qiao2020attention} and our Semantic Image Matting Dataset.}
\label{fig:classes_dist}
\vspace{-10pt}
\end{figure}

\begin{figure}[t]
\hspace{-0.1in}
\includegraphics[width=1.05\linewidth]{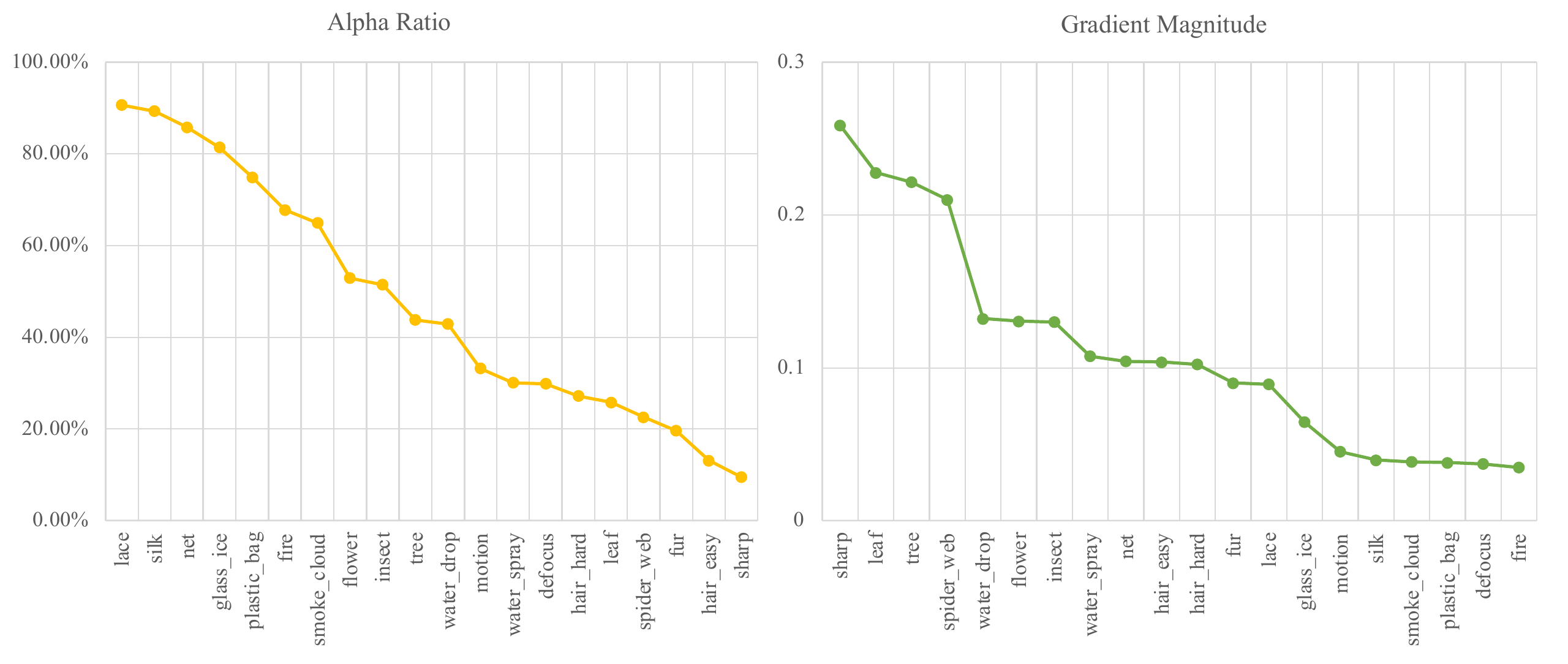} 
\caption{
Left: class-specific alpha ratio distribution. Alpha ratio is the quotient between the number of alpha pixels and the total number of all pixels within a region. Right: Alpha gradient magnitude distribution. Y-axis is the mean gradient magnitude of alpha pixels within a region.}
\label{fig:alpha_grad_hist}
\vspace{-10pt}
\end{figure}

\begin{figure*}[t]
\centering 
\includegraphics[width=1.0\linewidth]{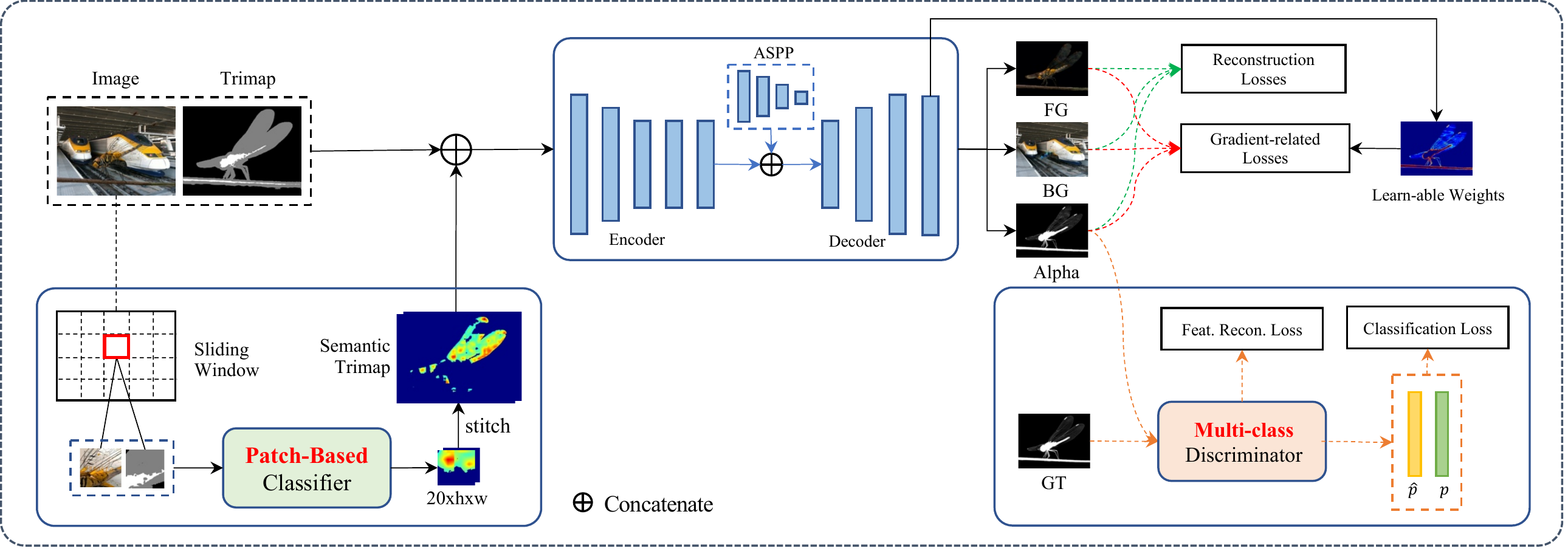} 
\caption{Overall framework. Our method takes as input an RGB image and its semantic trimap automatically generated from a patch-based classifier, which simultaneously predicts the alpha, foreground, background as well as learnable regularization weights. During training, a multi-class discriminator is used to provide supervision including classification loss and feature reconstruction loss from a semantic level.}
\label{fig:framework}
\vspace{-12pt}
\end{figure*}

\vspace{4pt}
\noindent{\textbf{Comparison with Other Matting Datasets.}} Although  large-scale matting datasets, e.g., Adobe Image Matting Dataset~\cite{Xu2017DeepIM} and Distinctions-646~\cite{qiao2020attention}, have been proposed as benchmarks in matting task, they are heavily biased toward common objects such as humans or animals.
Figure~\ref{fig:classes_dist} 
compares the statistics of these two representative datasets and our dataset, showing that  our Semantic Image Matting Dataset is class-balanced across the 20 matting classes.

\vspace{4pt}
\noindent{\textbf{Analysis.}} 
In Figure~\ref{fig:alpha_grad_hist}, 
 we compute the average gradient magnitude of alpha mattes as well as foreground alpha ratios within the unknown region in all classes and plot their distributions.  From the statistics, we can observe that while  some classes have smooth texture with small gradients, such as fire, smoke, and silk, other classes exhibit many fine features and sharp structures, such as net and spider web. This motivates us to exploit more useful information from class-related gradient constraints in training our model. More detailed statistics are included in the supplementary materials.
 
 Different matting classes exhibit quite distinctive alpha patterns and distributions, indicating that taking into consideration such semantics caused by different objects should boost matting performance.

\vspace{-2pt}
\section{Method}
\vspace{-2pt}
  Figure~\ref{fig:framework} shows 
our whole framework which is an encoder-decoder structure that takes an RGB image as well as its semantic trimap as input, and outputs the alpha predictions. Notably, this network is supervised by our multi-class discriminator as well as reconstruction and gradient-related losses. The network architecture of this multi-class discriminator is the same as the patch-based classifier which generates the semantic trimap, comprising of standard CNN, max-pooling layers, and ResBlocks.

\vspace{-1pt}
\subsection{Framework}
\label{sec:framework}

\vspace{2pt}
\noindent{\textbf{Patch-Based Classifier.}}
Semantic trimap is the concatenation of a conventional trimap and $n$-channel score maps where $n$ is the number of matting classes. The  conventional trimap defines definite foreground, definite background and unknown region while the score maps indicate the confidence each unknown pixel belongs to a certain matting class. The score maps are automatically generated through semi-supervised patch-based structure analysis within unknown regions. 

Our patch-based classifier is trained on the composited dataset where the images are partitioned with patch labels  indicating their respective matting classes. 
An alpha image usually contains more than one matting pattern, so it is partitioned into multiple regions each of which only belongs to a certain class. 
When training the classifier, we randomly crop a square patch  from unknown regions of a composited image. To make the classifier robust to scale variation, the crop size is random, uniformly chosen within a range of $[160,640]$ and then the cropped patch is resized to 320. Then, the classifier takes as input this given patch with the corresponding conventional trimap, and predicts the matting class for the given patch. 
After the classifier is well-trained, we compute the $n$-channel score maps for a patch by multiplying the last convolutional feature map with the fully connected weights, which is also known as class activation map~\cite{zhou2016cvpr}.

During inference, we obtain the semantic trimap for the entire input image through multi-scale patch analysis. In detail, we partition the input image into multi-scale overlapped patches and stitch the score maps of these patches together. Then we treat the stitched scores maps with the conventional trimap as our semantic trimap.

When we train our matting network, the semantic trimap will be concatenated with the RGB image and fed into the encoder, as shown in Figure~\ref{fig:framework}. Compared to conventional trimap, semantic trimap provides prior knowledge for the network to reduce search space for each class, focusing the model to predict more reliable alpha matte compatible to patterns caused by pertinent matting object semantics.

\vspace{4pt}
\noindent{\textbf{Encoder-Decoder Structure.}} Our framework for inference stage consists of U-Net~\cite{ronneberger2015u} like structure. The encoder is adapted from ResNet-50~\cite{he2016deep} with a downsampling stride of 8. Dilation convolution layers are used to enlarge receptive fields. Before sending encoded features to the decoder, an atrous spatial pyramid pooling (ASPP) module ~\cite{chen2017rethinking} is applied to aggregate features of different receptive fields in order to enhance the feature representation capability. Afterward, three up-conv layers are utilized to recover the spatial information by integrating high-level features as well as  high-resolution features from shallow layers. We simultaneously predict $F, B, \alpha$ through 3 prediction heads comprised of two $3\times3$ convolutional layers. 

\begin{figure}[t]
\centering 
\includegraphics[width=1.0\linewidth]{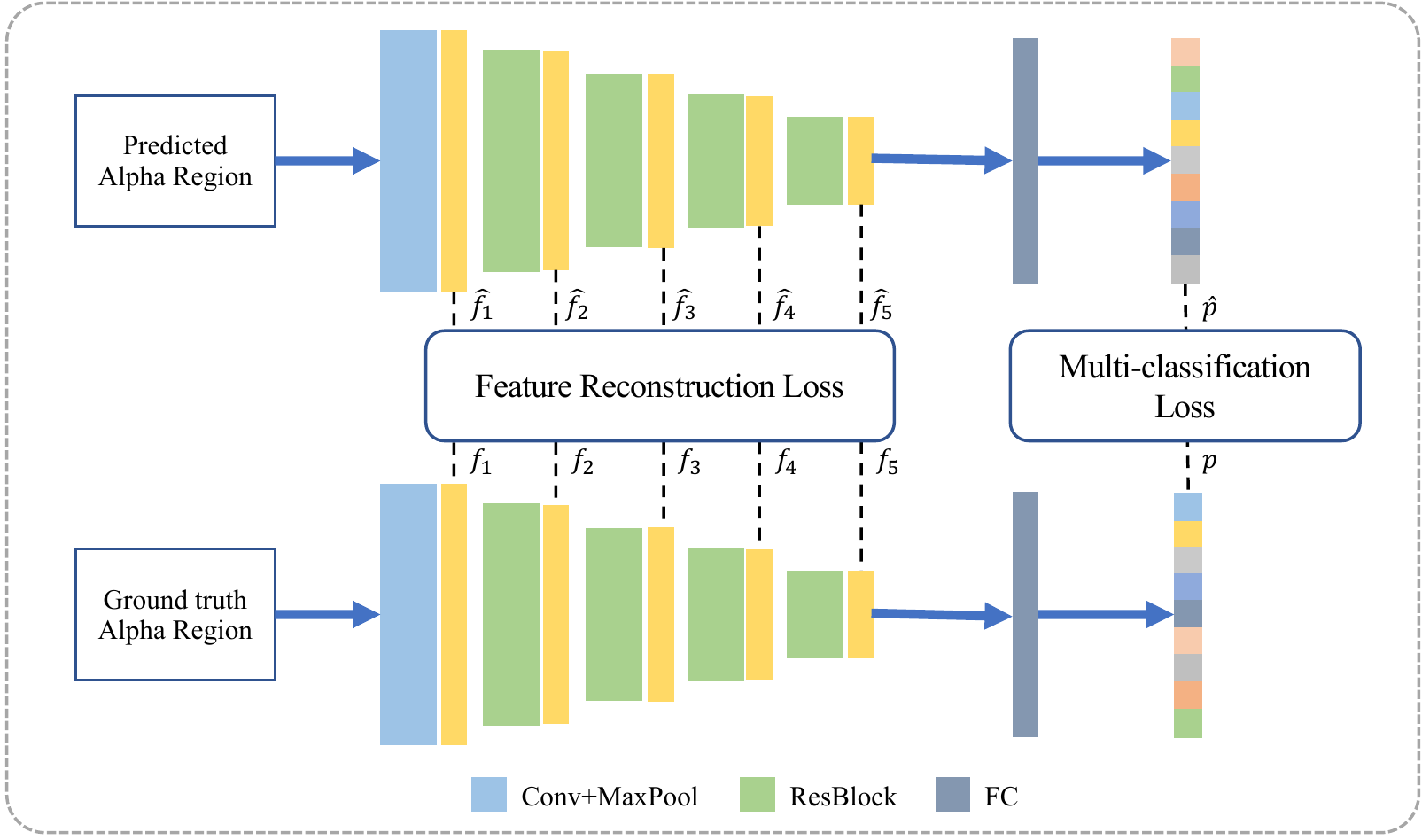} 
\caption{Multi-class discriminator. After feeding the predicted alpha region and groundtruth region to the well-trained discriminator, we compute the classification loss and feature reconstruction loss in order to preserve the distribution of predictions.}
\label{fig:discriminator}
\vspace{-\cspace pt}
\end{figure}

\vspace{4pt}
\noindent{\textbf{Multi-Class Discriminator.}}\label{sec:discriminator}
Previous works~\cite{Lutz2018AlphaGANGA, BMSengupta20, qiao2020attention} also adopt a discriminator to provide additional supervision for obtaining better alpha predictions. 
However, all of them have not taken inter-class difference into consideration. In contrast, we apply a multi-class discriminator to make the model aware of specific structures and preserve the statistics of relevant alpha patterns. 
The regularization imposed by the multi-class discriminator enforces the model to better learn the characteristics of each class and consequently give more reasonable predictions. 

Overall, the multi-class discriminator shares the same architecture with our semantic trimap classifier, except taking alpha matte as input during training and inference. Figure~\ref{fig:discriminator} shows its detailed structure with data flows. 

Before training the matting network, we first train this classifier on the groundtruth alpha mattes in Semantic Image Matting Dataset. 
The training process is similar to that of our semantic trimap classifier. Specifically, in each iteration, we randomly crop a square patch from the unknown class-specific regions and feed it to the classifier. After the classifier has been well trained, it will be deployed as a discriminator in our matting framework to generate classification loss and feature reconstruction loss.

During training, after obtaining the predicted alpha patch from the auto encoder-decoder matting network, the predicted alpha patch and its groundtruth patch are sent to the well-trained classifier so that we can extract for the two patches the respective multi-classification probabilities $\hat{p}$ and $p$, as well as the deep encoded multi-level features $\hat{f_k}$ and $f_k$ where $k\in\{1,2,3,4,5\}$. 

To compute the classification loss, we treat $\hat{p}$ as logit and $p$ as label, and apply a cross-entropy loss on them. Note that even though the prediction for the groundtruth patch $p$ is different from its original label, for instance, the original label is {\em hair\_easy} while the predicted label is {\em hair\_hard}, we still use $p$ as the label in the classification loss, since we suppose the distribution of the predicted alpha region should be as close as possible to that of the groundtruth alpha region.

To further improve the consistency, we  compute the feature reconstruction loss between the features of predicted alpha and groundtruth alpha, which is a part of the perceptual loss proposed in~\cite{johnson2016perceptual}. Conventionally, $L_2$ loss is performed between $\hat{f_k}$ and $f_k$ on each level separately.

\vspace {4pt}
\noindent{\textbf{Content-Sensitive Weights.}} Each matting class represents a distinct appearance and structure and thus its respective color and alpha exhibit different gradient distributions from other classes. For instance, {\em Hair} consists of fine structures with large gradients along  hair boundaries, while {\em Fire} exhibits smooth transition across its foreground region. 

Based on this observation, we introduce gradient constraints in our framework. 
The gradient of Eq.~\ref{eq:matting} is 
\begin{equation}\label{eq:gradient}
    \grad I = (F-B)\grad\alpha + 
    \alpha \grad F + (1-\alpha) \grad B
\end{equation}
Given an image, within the unknown region, $F-B$ and $\alpha$ are not available to the model. Thus, we learn 2D content-sensitive weights as different regularization coefficients to balance the gradient contributions among $F, B, \alpha$ for $I$. Specifically, the gradient constraint Eq.~\ref{eq:gradient} is re-formulated as Eq.~\ref{eq:grad_const} with 2D learnable weights $\lambda_1,\lambda_2$. 
\begin{equation}\label{eq:grad_const}
    \grad I = \lambda_1\grad\alpha + 
    (1-\lambda_2) \grad F + \lambda_2 \grad B
\end{equation}
By introducing gradient constraints with learnable content-sensitive weights, our framework learns the implicit relation between the gradient contribution and semantic representation, which guides the model to distinguish the source of the image structure. 

\begin{figure*}[ht]
\centering 
\includegraphics[width=1.0\linewidth]{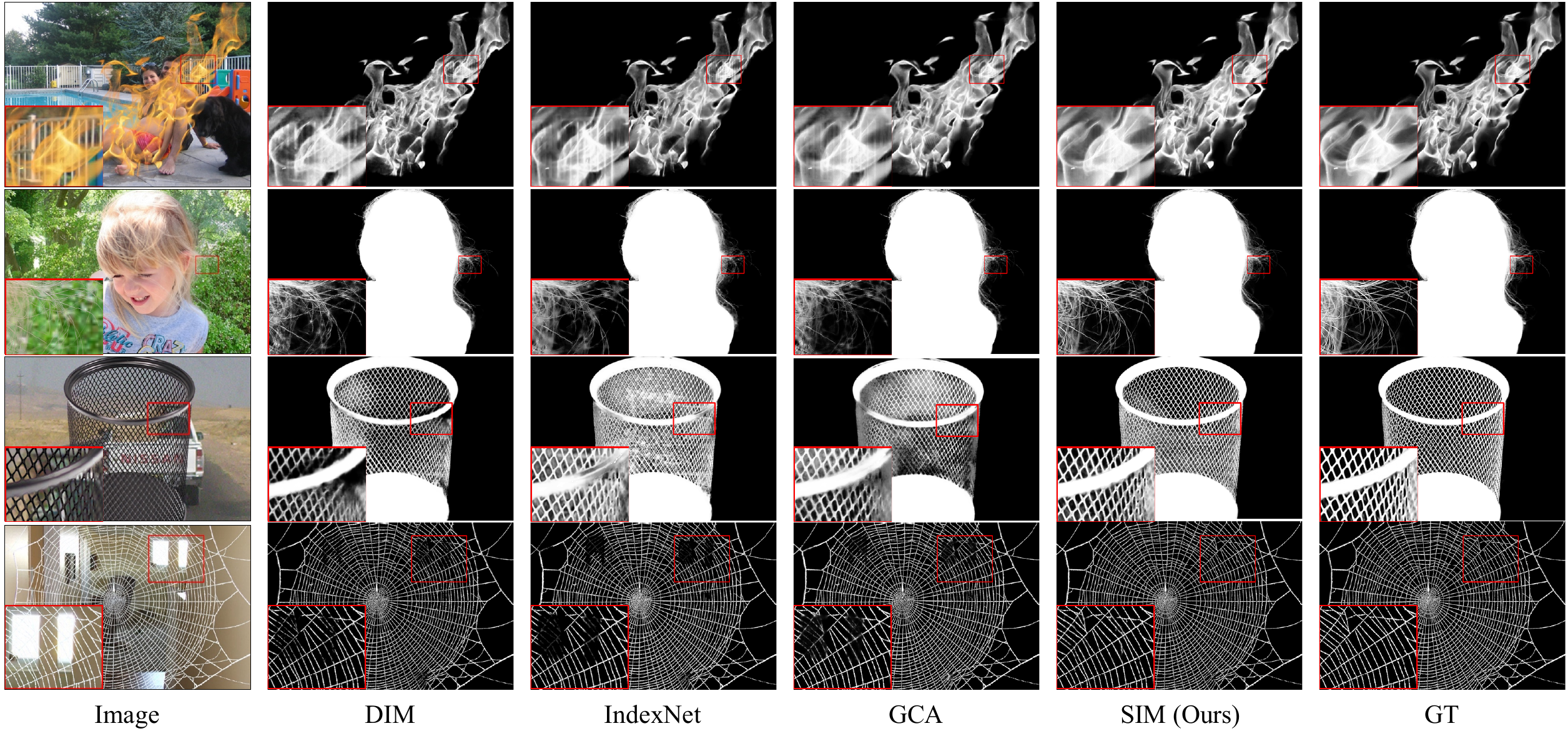} 
\caption{Qualitative results on the Semantic Image Matting test set.}
\label{fig:sim_res}
\vspace{-\cspace pt}
\end{figure*}

\subsection{Losses}
We jointly employ multiple losses, including reconstruction losses, classification loss, feature reconstruction loss, and gradient-related losses.

\vspace {4pt}
\noindent{\textbf{Reconstruction Losses.}} The reconstruction losses include alpha reconstruction loss $L_{\alpha}$ and $F,B$ reconstruction loss $L_{FB}$. Alpha reconstruction loss is composed of difference loss and composition loss as well as Laplacian pyramid loss $L_{lap}$ proposed by Hou~\etal\cite{hou2019context}, which is defined as Eq.~\ref{eq:loss1}. For foreground and background reconstruction loss, we only consider the pixels within $\widetilde{F}$ and $\widetilde{B}$ respectively, as $\widetilde{F}=F+U$ and $\widetilde{B}=B+U$:

\begin{equation}\label{eq:loss1}
    L_{\alpha} = {1\over{|U|}}\sum_{i\in U}\Vert\hat{\alpha}_i-\alpha_i\Vert_1
 + {1\over{|U|}}\sum_{i\in U}\Vert\hat{I}_i-I_i\Vert_1 + L_{lap}
 \end{equation}
\begin{equation}\label{eq:loss2}
    L_{FB} = {1\over{|\widetilde F|}}\sum_{i\in \widetilde F}\Vert\hat{F_i}-F_i\Vert_1 
+ {1\over{|\widetilde B|}}\sum_{i\in \widetilde B}\Vert\hat{B}_i-B_i\Vert_1
\end{equation}

\vspace {0pt}
\noindent{\textbf{Classification and 
Feature Reconstruction Loss.}} 
As discussed above, our multi-class discriminator regularizes the model from a semantic level by classification loss $L_c$ and feature reconstruction loss $L_f$ defined as below:
\begin{equation}\label{eq:loss3}
    L_c = -\sum_{j}\hat{p}_j\log{p_j}
\end{equation}
\begin{equation}\label{eq:loss3_1}
L_f = \sum_{k}{1\over{|f_k|}}\Vert\hat{f}_k-f_k\Vert_2
\end{equation}

\vspace {0pt}
\noindent{\textbf{Gradient-related Loss.}} To enforce the model to obey the gradient constraint, we utilize the loss with learnable content-sensitive regularization weights as Eq.~\ref{eq:loss4}:
\begin{equation}\label{eq:loss4}
    L_g = {1\over{|U|}}\sum_{i\in U}\Vert\grad \hat{I}_i- \grad I_i\Vert_1 
\end{equation}
\begin{equation}
   \grad\hat{I}_i = \lambda_1\grad\hat{\alpha}_i + 
    (1-\lambda_2) \grad\hat{F}_i + \lambda_2 \grad\hat{B}_i
\end{equation}

In addition, to fully exploit the gradient constraint, an exclusion loss used in a similar form in~\cite{exclusiveloss} is defined in Eq.~\ref{eq:loss5} to avoid the structure of the image leaking into both foreground and background: 
\begin{equation}\label{eq:loss5}
    L_e = {1\over{|U|}}\sum_{i\in U}\Vert\grad\hat{F}_i\Vert_1\Vert\grad\hat{B}_i\Vert_1 + \Vert\grad\hat{\alpha}_i\Vert_1\Vert\grad\hat{B}_i\Vert_1
\end{equation}

\vspace {4pt}
\noindent{\textbf{Total Loss.}} The total loss is thus the weighted sum of the above losses defined as: 
\begin{equation}\label{eq:loss6}
    L = L_{\alpha} + 0.2(L_{FB}  + L_f + L_g + L_e) + 0.1L_c
\end{equation}

\section{Experiments}

\begin{table}[t]
    \centering
    \setlength{\tabcolsep}{2.2mm}{
    \begin{tabular}{p{2cm}|cccc}
        \hline\hline
        Methods & SAD & MSE($10^3$) & Grad & Conn \\
        \hline
        DIM~\cite{Xu2017DeepIM} & 48.07 & 15.0 & 31.67 & 46.26 \\
        IndexNet~\cite{lu2019indices} & 51.29 & 14.0 & 34.19 & 48.77 \\
        GCA~\cite{li2020natural} & 39.28 & 11.0 & 28.70 & 36.03 \\
        SIM (Ours) & \textbf{27.87} & \textbf{4.7} & \textbf{11.57} & \textbf{20.83} \\
        \hline\hline       
    \end{tabular}}
    \vspace{5pt}
    \caption{Comparisons on Semantic Image Matting Dataset.}
    \label{tab:comparison}
    \vspace{-\cspace pt}
\end{table}

\begin{figure*}[t]
\centering 
\includegraphics[width=1.0\linewidth]{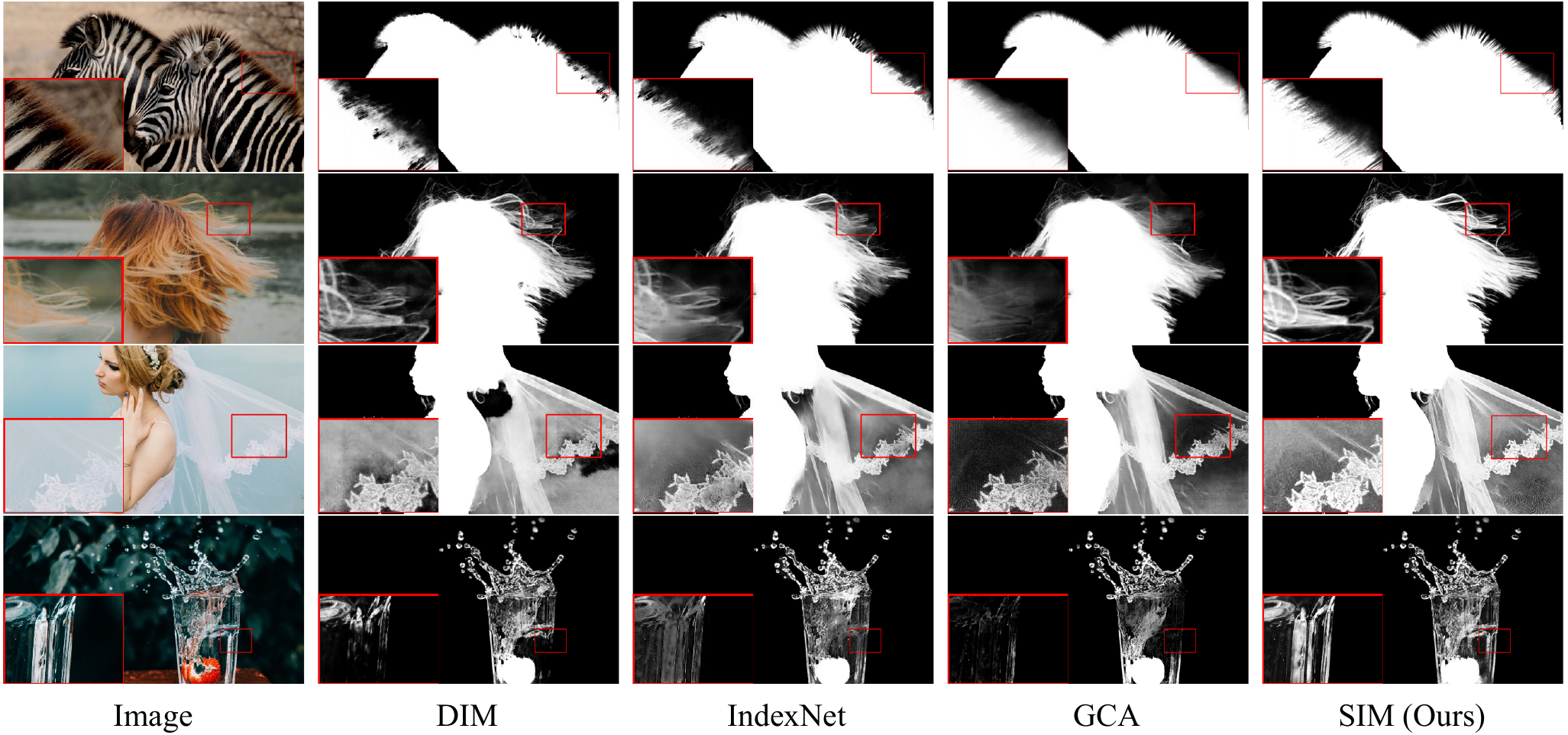} 
\caption{Qualitative results on the real-world images.}
\label{fig:sim_real_res}
\vspace{-\cspace pt}
\end{figure*}

\subsection{Implementation Details}

We train our models on the Semantic Image Matting Dataset. Specifically, we take the images in COCO dataset~\cite{lin2014microsoft} as background images. For each foreground object, we randomly select an image from the backgrounds and composite them using Eq.~\ref{eq:matting}. Before composition, we perform various augmentations on both foreground and background. For the foreground, we randomly crop a square patch from size $[320,480,640]$ and apply random scaling, rotation, color jittering, and horizontal flipping on the patch. Meanwhile, we randomly crop a patch from the background image and then generate inputs from the two patches. To extend training samples, we also randomly merge two foregrounds into one following~\cite{li2020natural}. 
We dilate and erode alpha matte with a random kernel size within $[2,9]$ and a random iteration within $[5, 15]$ for trimap generation. More details can be found in  supplementary materials. 

\begin{table}[t]
    \centering
    \setlength{\tabcolsep}{1.5mm}{
    \begin{tabular}{l|cccc}
        \hline\hline
        Methods & SAD & MSE($10^3$) & Grad & Conn \\
        \hline\hline
        Closed-Form~\cite{levin2008closed} &  168.1 & 91.0 & 126.9 & 167.9  \\
        KNN-Matting~\cite{Chen-2012-PAMI-knnmatting} & 175.4 & 103.0 & 124.1 & 176.4 \\
        DCNN-Matting~\cite{Cho-2016-mattingusingdeepcnn} & 161.4 & 87.0 & 115.1 & 161.9 \\
        Learning-Based~\cite{zheng2009learning} & 113.9 & 48.0 & 91.6 & 122.2 \\
        Information-flow~\cite{Aksoy-2017-cvpr-ifm} & 75.4 & 66.0 & 63.0 & - \\
        DIM~\cite{Xu2017DeepIM} & 50.4 & 14.0 & 31.0 & 50.8 \\
        AlphaGan~\cite{Lutz2018AlphaGANGA} & 52.4 & 30.0 & 38.0 & - \\
        IndexNet~\cite{lu2019indices} & 45.8 & 13.0 & 25.9 & 43.7 \\
        AdaMatting~\cite{cai2019disentangled} & 41.7 & 10.0 & 16.9 & - \\
        SampleNet~\cite{tang2019learning} & 40.4 & 9.9 & - & - \\
        Context-Aware~\cite{hou2019context} & 35.8 & 8.2 & 17.3 & 33.2 \\
        GCA~\cite{li2020natural} & 35.3 & 9.1 & 16.9 & 32.5 \\
        SIM (Ours) & \textbf{28.0} & \textbf{5.8} & \textbf{10.8} & \textbf{24.8} \\
        \hline\hline       
    \end{tabular}}
    \vspace{2pt}
    \caption{Quantitative results on Composition-1K ~\cite{Xu2017DeepIM}.}
    \label{tab:composition1k}
    \vspace{- \cspace pt}
\end{table}

\vspace{-1pt}
\subsection{Main Results}
\vspace{\cspaceadd pt}
\vspace{-2pt}
\noindent{\textbf{Results on Semantic Image Matting Dataset.}} 
Table~\ref{tab:comparison} tabulates the quantitative comparisons with other methods on our Semantic Image Matting Dataset. 
Our method outperforms them and achieves state-of-the-art performance. Table~\ref{tab:classes} further lists the quantitative results on each matting class. Our method also performs better than other methods on all matting classes. Figure~\ref{fig:sim_res} show the qualitative comparisons of some classes on the test set.

\vspace{\cspaceadd pt}
\noindent{\textbf{Results on Composition-1K.}} We also evaluate our model on the Composition-1K dataset proposed by Xu et al.~\cite{Xu2017DeepIM}, which has 50 unique foreground objects and 1000 testing samples. 
 Table~\ref{tab:composition1k} shows the quantitative results. Our method also achieves the most competitive state-of-the-art performance on this dataset. 

\vspace{\cspaceadd pt}
\noindent{\textbf{Results on alphamatting.com.}} Table~\ref{tab:alphamatting} shows the evaluation results on alphamatting.com~\cite{alphamatting} benchmark proposed in ~\cite{Rhemann-2009-perceptuallybenchmarkimage}. Our method outperforms other methods and ranks first on three metrics. Detailed comparisons are provided in supplementary materials.

\begin{table*}[t]\small
    \centering
    \setlength{\tabcolsep}{1.6mm}{
    \begin{tabular}{l|c|c|c|c|c|c|c|c|c|c}
        \hline\hline
        Classes & defocus & fire & fur & glass\_ice & hair\_easy & hair\_hard & insect & motion & net & flower \\ 
        \hline
        DIM~\cite{Xu2017DeepIM} & 25.91 & 60.53 & 9.88 & 91.36 & 11.23 & 13.01 & 111.21 & 6.78 & 87.09 & 65.40 \\
        IndexNet~\cite{lu2019indices} & 22.86 & 97.85 & 9.99 & 91.95 & 8.33 & 13.24 & 130.52 & 6.68 & 91.43 & 59.60 \\
        GCA~\cite{li2020natural} & 18.33 & 46.29 & 8.12 & 76.20 & 8.24 & 11.31 & 99.11 & 6.08 & 83.71 & 44.86  \\
        SIM (Ours) & \textbf{13.49} & \textbf{35.44} & \textbf{5.90} & \textbf{49.19} & \textbf{5.68} & \textbf{7.72} & \textbf{96.85} & \textbf{4.04} & \textbf{50.35} & \textbf{37.10} \\ 
        \hline\hline
        Classes & leaf & tree & plastic\_bag & sharp & smoke\_cloud & spider\_web & lace & silk & water\_drop & water\_spray \\
        \hline
        DIM~\cite{Xu2017DeepIM} & 45.43 & 91.71 & 65.44 & 2.96 & 48.21 & 145.57 & 101.78 & 51.89 & 32.48 & 41.96 \\
        IndexNet~\cite{lu2019indices} & 43.85 & 99.26 & 89.70 & 3.32 & 35.31 & 145.62 & 114.47 & 62.81 & 33.90 & 34.92 \\
        GCA~\cite{li2020natural} & 41.12 & 87.61 & 47.40 & 3.35 & 41.18 & 107.14 & 80.51 & 51.93 & 25.83 & 31.12 \\
        SIM (Ours) & \textbf{20.98} & \textbf{34.14} & \textbf{36.70} & \textbf{1.39} & \textbf{27.42} & \textbf{63.79} & \textbf{51.08} & \textbf{41.78} & \textbf{16.94} & \textbf{20.53} \\
        \hline\hline       
    \end{tabular}}
    \vspace{5pt}
    \caption{Quantitative results of 20 classes on Semantic Image Matting Dataset.}
    \label{tab:classes}
    \vspace{-10pt}
\end{table*}

 \begin{table}[t]\small
    \centering
    \setlength{\tabcolsep}{1mm}{
    \begin{tabular}{l|cccc|c|c}
    \hline\hline
    \multicolumn{1}{c|}{\multirow{2}{*}{Methods}} &
    \multicolumn{4}{c|}{\multirow{1}{*}{SAD}} &
    \multicolumn{1}{c|}{\multirow{1}{*}{MSE}} &
    \multicolumn{1}{c}{\multirow{1}{*}{Grad}} \\
    \cline{2-7}
    \multicolumn{0}{c|}{} &
    \multicolumn{1}{c}{\multirow{1}{*}{Overall}} &
    \multicolumn{1}{c}{\multirow{1}{*}{S}} &
    \multicolumn{1}{c}{\multirow{1}{*}{L}} &
    \multicolumn{1}{c|}{\multirow{1}{*}{U}} &
    \multicolumn{1}{c|}{\multirow{1}{*}{Overall}} &
    \multicolumn{1}{c}{\multirow{1}{*}{Overall}} \\
    \hline
    AdaMatting~\cite{cai2019disentangled} & 7.6 & 6.9 & 6.5 & 9.4 & 8.5  & 8.1 \\
    SampleNet~\cite{tang2019learning} & 8.2 & 6.5 & 7.6 & 10.5 & 9.2 & 9.5 \\
    Background~\cite{BMSengupta20} & 7.9 & 5.9 & 5.4 & 12.4 & 7.4  & 6.9 \\
    GCA~\cite{li2020natural} & 9 & 10 & 6.4 & 10.8 & 9.9 & 8.2  \\
    SIM (Ours) & \textbf{2.5} & \textbf{2.6} & \textbf{1.8} & \textbf{3}  & \textbf{2.9} & \textbf{3.1}  \\
    \hline\hline       
    \end{tabular}}
    \vspace{3pt}
    \caption{Quantitative results of our method and several representative state-of-the-art methods on alphamatting.com~\cite{alphamatting} benchmark. ``S'', ``L'', ``U'' denote three trimap sizes and scores denote average rank across 8 test samples. Best results are shown in bold.}
    \label{tab:alphamatting}
    \vspace{-5pt}
\end{table}

\vspace{\cspaceadd pt}
\noindent{\textbf{Results on Real-World Images.}} To evaluate the generalization ability of our framework, we collect from the Internet a number of  free real-world images of different matting classes with user-labeled trimaps.  Figure~\ref{fig:sim_real_res} shows the qualitative results which further demonstrates the effectiveness and generalization ability of our framework.

\begin{table}[t]
    \centering
    \setlength{\tabcolsep}{1.8mm}{
    \begin{tabular}{p{2.6cm}|cccc}
        \hline\hline
        Methods & SAD & MSE($10^3$) & Grad & Conn \\
        \hline
        Basic & 32.04 & 5.9 & 12.05 & 26.20 \\
        Basic + S & 30.24 & 5.4 & 11.60 & 23.83 \\
        Basic + S + D & 29.84 & 5.3 & 12.37 & 23.33 \\
        Basic + S + D + G & 27.87 & 4.7 & 11.57 & 20.83 \\
        \hline\hline
    \end{tabular}}
    \vspace{1pt}
    \caption{Ablation studies. The basic model is trained with reconstruction losses. ``S", ``D", ``G" denotes semantic trimap , multi-class discriminator and gradient-related losses respectively.}
    \label{tab:submodule}
    \vspace{-\cspace pt}
    \vspace{1pt}
\end{table}

\vspace{-2pt}
\subsection{Analysis}
\begin{figure}[t]
\centering 
\includegraphics[width=1.0\linewidth]{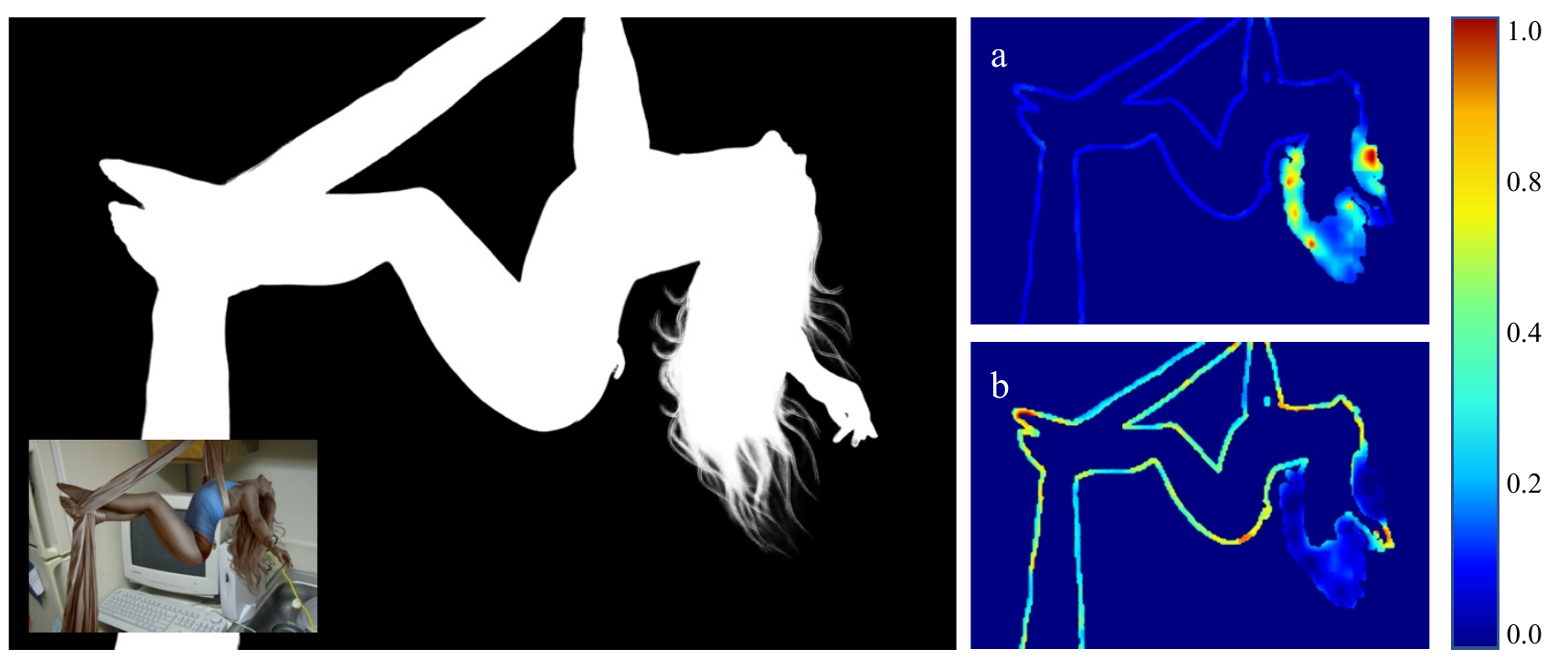} 
\caption{An example semantic trimap. We visualize two channels of the score maps: $a$. $hair\_hard$; $b$. $sharp$.}
\label{fig:smap}
\vspace{-4pt}
\end{figure}

\vspace{\cspaceadd pt}
\vspace{-2pt}
\noindent{\textbf{Analysis of Semantic Trimap.}} 
Figure~\ref{fig:smap} visualizes an example semantic trimap, which is the combination of a conventional trimap and an automatically generated 20-channel score maps. The conventional trimap is divided into foreground, background, and unknown regions, which does not provide any semantics of the unknown pixels. Although deep neural networks are capable of encoding the implicit semantics of alpha patterns into high-level features, many fail cases still cannot be avoided due to restricted receptive fields and complex background, especially when the conventional trimap is too coarse to provide enough clues for extracting the relevant foreground object. The semantics score maps on the other hand provide the much-needed prior knowledge for indicating the potential class of unknown pixels, which guides the network to generate more accurate predictions. 

\vspace{\cspaceadd pt}
\vspace{4pt}

\noindent{\textbf{Analysis of Multi-Class Discriminator.}} 
The arguably most widely used criterion in matting optimization is   pixel-to-pixel distance which unfortunately does not take any distribution of a region into consideration. Thus, we propose a  multi-class discriminator to make the model learn the distribution of different classes by taking into consideration class-specific statistics of alpha mattes.

Previous works~\cite{torralba2003statistics, mechrez2018contextual} had exploited natural image statistics and utilized them to boost performance in image reconstruction or translation. As for matting, if the predicted mattes are satisfactory, such predictions are supposed to have similar distributions to those of the groundtruth. The classification and feature reconstruction losses provided by the multi-class discriminator enforces the model to preserve the statistics of different patterns and consequently improves the performance. 

\begin{figure}[t]
\centering 
\includegraphics[width=1.0\linewidth]{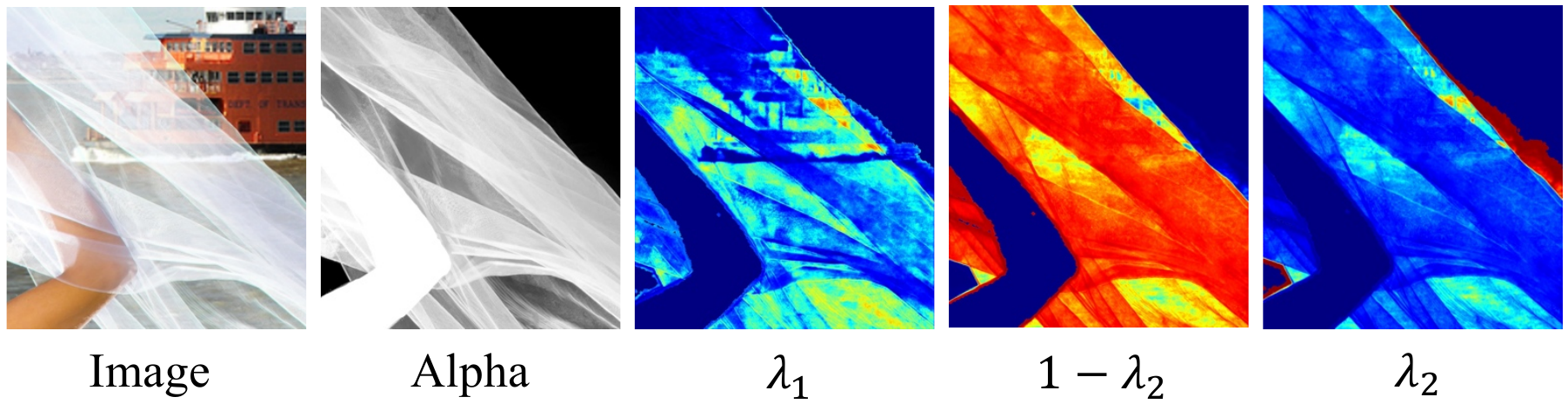} 
\caption{Visualization of learnable weights.} 
\label{fig:attention}
\vspace{-\cspace pt}
\end{figure}

\vspace{\cspaceadd pt}
\vspace{4pt}
\noindent{\textbf{Analysis of Content-Sensitive Weights.}} 
During the training stage, we introduce the gradient constraints with learnable weights to regularize the model on different classes.  Figure~\ref{fig:attention} visualizes sample learnable weights. 
From this example, we observe that for the semi-transparent smooth textures such as silk, the alpha matte and the background contribute most to the image gradient. 
With the learnable content-sensitive weights, the performance is improved by 1.97 on SAD as shown in Table~\ref{tab:submodule}.

\section{Conclusion}
\vspace{-8pt}
In this paper, we make use the semantics underlying different alpha mattes, and develop a method incorporating semantic classification of matting regions for extracting better alpha matte. 
Specifically, we first cluster 20 classes according to regional alpha patterns, which cover a wide range of matting scenarios. Based on the classes, we extend traditional trimap to semantic trimap which is automatically extracted by a classifier to provide valuable semantic information for predictions. 
To further preserve alpha statistics for each class, a novel multi-class discriminator is designed to regularize the model according to class-specific distributions. Finally, we introduce an alpha gradient constraint with learnable content-sensitive weights as a new regularization to achieve better optimization of different classes. We conduct extensive experiments on Semantic Image Matting Dataset, Composition-1K dataset, and real-world images. 
Quantitative and qualitative results demonstrate the clear advantages of our method over existing methods.

{\small
\bibliographystyle{ieee_fullname}
\bibliography{egbib}
}

\end{document}